\documentclass[10pt,twocolumn,letterpaper]{article}

\usepackage{iccv}
\usepackage{times}
\usepackage{epsfig}
\usepackage{graphicx}
\usepackage{amsmath}
\usepackage{amssymb}
\usepackage{epstopdf}
\usepackage{enumerate}
\usepackage{graphicx} 
\usepackage{subfigure}
\usepackage{tikz}
\usepackage{pgfplots}
\usepackage{pgflibraryarrows}
\usepackage{pgflibrarysnakes}
\usepackage{multirow}
\usepackage{amsmath,bm}
\usepackage[eulergreek]{sansmath}
\usepackage[latin1]{inputenc}
\usepackage{helvet}
\usepackage{color}
\usepackage{pifont}
\usepackage{makecell}
\usepackage{verbatim}
\newcommand{\cmark}{\ding{51}}%
\newcommand{\xmark}{\ding{55}}%
\usepackage{tikz}
\usepackage{etoolbox}
\usepackage{authblk}
\newcommand{\circled}[2][]{\tikz[baseline=(char.base)]
    {\node[shape = circle, draw, inner sep = 0.5pt]
    (char) {\phantom{\ifblank{#1}{#2}{#1}}};%
    \node at (char.center) {\makebox[0pt][c]{#2}};}}
\robustify{\circled}

\usepackage[pagebackref=true,breaklinks=true,letterpaper=true,colorlinks,bookmarks=false]{hyperref}

\iccvfinalcopy 

\def\iccvPaperID{****} 

\ificcvfinal\fi

\begin{document}

\title{Progressive Sparse Local Attention for Video Object Detection}

\author{Chaoxu Guo$^{1,2}$\quad Bin Fan$^{1}$\thanks{Bin Fan is the corresponding author}\quad Jie Gu${}^{1}$\quad Qian Zhang$^{3}$ \quad Shiming Xiang${}^{1,2}$ \\
\vspace{-0.4cm}
  \quad V\'eronique Prinet${}^{1}$ \quad Chunhong Pan${}^{1}$\\
${}^{1}$National Laboratory of Pattern Recognition, Institute of Automation, Chinese Academy of Sciences\\
${}^{2}$School of Artificial Intelligence, University of Chinese Academy of Sciences\\
${}^{3}$Horizon Robotics  \\
\emph{\{chaoxu.guo,bfan,smxiang,prinet,chpan\}@nlpr.ia.ac.cn},
\emph{\{qian01.zhang\}@horizon.ai}
}

\maketitle
\ificcvfinal\thispagestyle{empty}\fi

\begin{abstract}
   Transferring image-based object detectors to the domain of videos remains a challenging problem. Previous efforts mostly exploit optical flow to propagate features across frames, aiming to achieve a good trade-off between accuracy and efficiency. However, introducing an extra model to estimate optical flow can significantly increase the overall model size. The gap between optical flow and high-level features can also hinder it from establishing spatial correspondence accurately. Instead of relying on optical flow, this paper proposes a novel module called Progressive Sparse Local Attention (PSLA), which establishes the spatial correspondence between features across frames in a local region with progressively sparser stride and uses the correspondence to propagate features. Based on PSLA, Recursive Feature Updating (RFU) and Dense Feature Transforming (DenseFT) are proposed to model temporal appearance and enrich feature representation respectively in a novel video object detection framework. Experiments on ImageNet VID show that our method achieves the best accuracy compared to existing methods with smaller model size and acceptable runtime speed.
\end{abstract}

\section{Introduction}

    Object detection is a fundamental problem in computer vision and serves as a core technique in many practical applications, \eg robotics, autonomous driving and human behavior analysis. With the development of convolutional neural networks (CNNs), remarkable successes have been achieved on detecting objects from images~\cite{rfcn,fast-rcnn,rcnn,mask-rcnn, spp,fpn,ssd,yolo,faster-rcnn}. However, applying those techniques on a frame-by-frame basis to a video is often unsatisfactory due to the deteriorated appearance caused by issues such as motion blur, out-of-focus camera, and rare poses frequently encountered in videos. The temporal information encoded inherently in videos has been used to improve the performance of video object detection, as it provides rich cues about the motion in videos that are absent from still images.

    \begin{figure}
    \centering
    \includegraphics[width=0.5\textwidth]{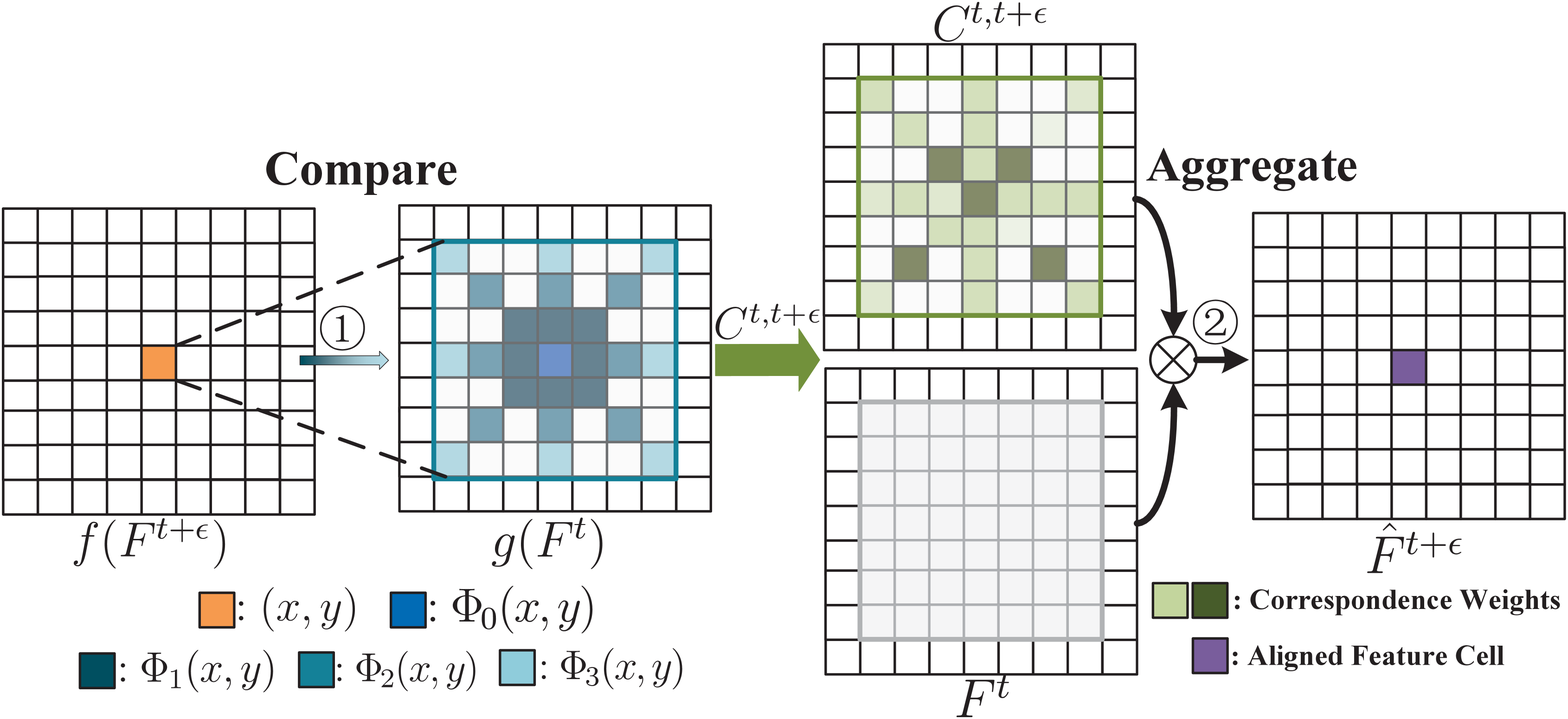}
    \caption{\footnotesize Illustration of Progressive Sparse Local Attention (PSLA). The goal of PSLA is to align feature map $F_{t}$ with $F^{t+\epsilon}$ in an attention way, which is formulated into two steps: the first step $\circled{1}$ is that each feature cell in embedded feature map $f(F^{t+\epsilon})$ compares to the surrounding cells in the embedded feature map $g(F^t)$, in a progressive sparser stride from center to outside. Regions with different colors in $g(F^t)$ represent regions in different strides, which are illustrated in equation \ref{eq2} and \ref{eq3}. The resulting feature affinities are used to compute correspondence weights $C^{t,t+\epsilon}$, which capture the spatial correspondence between features. The second step $\circled{2}$ is that the chosen feature cells in $F^t$ are aggregated with the corresponding weights to generate a feature cell in $\hat{F}^{t+\epsilon}$, which is the aligned feature map from $F^t$.}
    \label{fig1}
    \end{figure}

    Existing methods that leverage temporal information for object detection from videos mainly fall into two categories. The first one relies on dedicated post-processing \cite{seqnms,tcnn,kang2016object,lee2016multi}. These methods firstly run an image-based detector on single frames and then integrate the per-frame results by box-level post-processing, which usually requires an extra object tracker or off-the-shelf optical flow to estimate the motion field and associate the bounding boxes. Modeling temporal coherence in this way is sub-optimal since detectors do not benefit from the temporal information in the phase of training.

    Another category of methods \cite{lattice,DT, fully,stmn,towards,fgfa,dff} exploits the temporal information in videos when training the detectors. They either pursue a trade-off between accuracy and complexity or seek to improve performance at the expense of runtime. Among these methods, the optical flow is widely used to propagate the high-level features across frames. Extra optical flow models, \textit{e.g.} FlowNet \cite{flownet}, have to be utilized to enable the end-to-end training and achieve better performance. However, adding an optical flow model has several drawbacks. First, the extra model significantly increases the overall model size of detectors  (\textit{e.g.}, a typical detector of ResNet101+RFCN has 59.6M parameters and it has to add additional 37M parameters when using FlowNet.), which makes it harder to be deployed on mobile devices. Second, optical flow only establishes local pixel correspondences between two images. Directly transferring the flow field to high-level features may introduce artifacts because it ignores the transformation that happens from layer to layer in the network. Finally, a shift of one pixel in high-level feature maps may correspond to up to tens of pixels in the image. It is very challenging for optical flow to capture such a large displacement.

    Our work belongs to the second category. To address the limitations above, we propose a novel module, Progressive Sparse Local Attention (PSLA), to propagate high-level semantic features across frames without relying on optical flow. Specifically, given two features $F^t$ and $F^{t+\epsilon}$ of frames $I^t$ and $I^{t+\epsilon}$ respectively, PSLA first produces correspondence weights based on the feature affinities between $F^t$ and $F^{t+\epsilon}$ and then aligns $F^t$ with $F^{t+\epsilon}$ by aggregating features with corresponding weights. It is similar to attention mechanisms \cite{attention} but different in that the attended positions in PSLA are distributed in a local region with progressive sparser strides as illustrated in Fig. \ref{fig1}, which is inspired by the motion distribution in videos as shown in Fig. \ref{fig3}.

    Based on PSLA, a video object detection framework is proposed, in which expensive extraction of high-level features is performed on sparse key frames while low-cost extraction of low-level features is applied on dense non-key frames. Based on the extracted features, PSLA is used in two distinct and complementary situations: (1) to propagate high-level features (at a given layer of the network) from key frames to non-key frames. This allows us to assign most of the computation cost to key frames and improves efficiency when testing without sacrificing accuracy. Moreover, a small network named \textit{Quality Net} is devised to complement the propagated high-level features with low-level information from features of non-key frames, in order to reduce the aliasing effect of feature propagation. We name this procedure Dense Feature Transforming (DenseFT). (2) to maintain a temporal feature $F_t$ that models temporal appearance of the video, by propagating high-level features across key frames. Meanwhile, an \textit{Update Net} is proposed to recursively update $F_t$ with high-level features of key frames. Our ablation study shows that exploiting temporal context contributes to a substantial gain in performance. We name this procedure Recursive Feature Updating (RFU).

    We conducted extensive experiments on ImageNet VID \cite{imagenet} for video object detection. Our results are on par with or outperform state-of-the-art methods in both speed and accuracy with reduced model size. In addition, we show that our model can generalize to other tasks such as video semantic segmentation on the CityScapes dataset \cite{cityscapes}.

    In summary, the contributions of this paper include:
    \begin{itemize}
        \item We propose a novel module Progressive Sparse Local Attention (PSLA) to establish the spatial correspondence between feature maps without relying on extra optical flow models, which reduces model parameters significantly while achieves better results.
        \vspace{-0.2cm}
        \item Based on PSLA, two techniques, Recursive Feature Updating (RFU) and Dense Feature Transforming (DenseFT), are developed to model temporal appearance and enhance the feature representation of non-key frames, respectively.
        \vspace{-0.2cm}
        \item We introduce a novel framework for video object detection which achieves state-of-the-art performance on ImageNet VID \cite{imagenet}.
    \end{itemize}

\section{Related Work}

    {\bf Image Object Detection.}
    Existing state-of-the-art methods for image object detection mostly follow two paradigms, two-stage and single-stage. A two-stage pipeline consists of generation of region proposals, region classification, and location refinement. R-CNN \cite{rcnn} is a seminal work of two-stage methods. Fast R-CNN \cite{fast-rcnn} improves the speed and accuracy by sharing computation of feature extraction while Faster R-CNN\cite{faster-rcnn} learns to generate region proposals. Some following variants, \textit{e.g.} R-FCN \cite{rfcn} and FPN \cite{fpn}, further improve the performance. Comparing to two-stage detectors, single-stage methods are more efficient but less accurate. SSD \cite{ssd} produces detection results from default anchor boxes from multiple feature maps. YOLO \cite{yolo,yolov2} formulate detection as a regression problem. Lin \textit{et al.} \cite{focal} propose focal loss to address the problem of data imbalance. In this paper, we use R-FCN as our base detector.

    {\bf Video Object Detection.}
    Different from image object detection, methods for video detection should take temporal information into account. T-cnn \cite{tcnn} utilizes off-the-shelf optical flow to propagate bounding boxes. Then the boxes are re-scored and removed by considering the temporal context of videos. Tpn \cite{tpn} proposes a tubelet proposal network and employs an LSTM to incorporate temporal information from tubelet proposals. To boost the performance, MANet \cite{fully} and FGFA \cite{fgfa} use the optical flow estimated by FlowNet \cite{flownet} to aggregate the feature of multiple nearby frames. Instead of relying on optical flow, D\&T \cite{DT} predicts the bounding boxes of the next frame by performing correlation between the features of current and next frame. To reduce the computation cost, Zhu \etal \cite{dff,towards} use optical flow to propagate the high-level features of key-frames to other frames and avoid extracting expensive feature frequently. Chen \etal \cite{lattice} improve both speed and accuracy by designing a time-scale lattice. But an extra classifier is required to re-score the bounding boxes. It increases the model parameters greatly. The closest work to ours is STMN \cite{stmn}, which utilizes a module similar to correlation to align feature maps in a local region. Different from~\cite{stmn}, our approach focuses on a sparse neighborhood and \text{softmax} normalization is utilized to better establish spatial correspondence. We improves both speed and accuracy while STMN improves accuracy at the expense of runtime.

    {\bf Self-attention.}
    Self-attention is a mechanism first introduced in \cite{attention} for machine translation. To integrate enough context and long-range information in a sequence, it computes the response at a position in a sequence by taking the weighted mean values of all positions, where the weights are learned by backpropagation without explicit supervision. Bahdanau \etal \cite{bahdanau2014neural} apply soft attention to machine translation aiming to capture soft alignments between the source and target words. Unlike those previous works \cite{bahdanau2014neural,attention}, our proposed PSLA is a more general form of self-attention. In this paper, it is applied in the temporal-spatial domain towards aligning two feature maps.

    {\bf Nonlocal Operators.}
    Nonlocal is a traditional filter algorithm \cite{buades2005non} that is widely used in image denoising \cite{image-denoise1,image-denoise2}, super-resolution \cite{super} and texture synthesis \cite{texture}. Those approaches compute response as a weighted mean of all the pixels in an image, where the weights are obtained based on the patch appearance similarity. More recently, based on the same principle, Wang \etal \cite{nonlocal} develop a nonlocal operator utilized for video classification and object detection. It aims to capture long-range dependency within feature maps and augments the receptive field. This operator is further extended to image generation \cite{sagan} and semantic segmentation \cite{danet,ccnet,ocnet}. Different from those methods, PSLA focuses on a local region with progressive sparser strides.

\section{The Proposed Method}
\subsection{Overview}\label{framework}
    \begin{figure}
    \centering
    \includegraphics[width=0.5\textwidth]{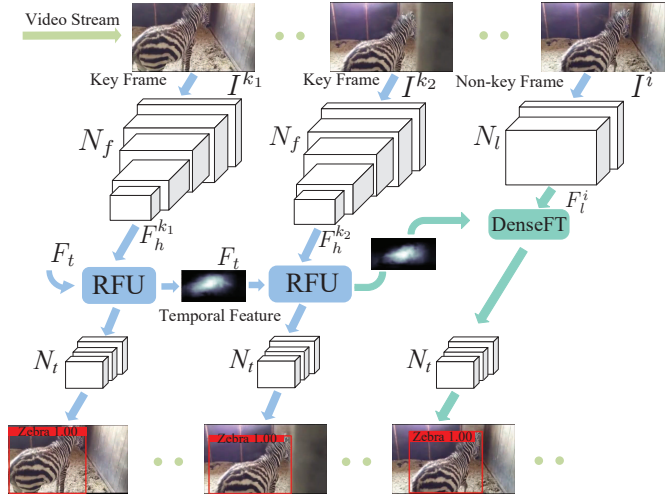}
    \caption{\footnotesize Pipeline of the proposed video detection framework. Only two key frames $I^{k1}$, $I^{k2}$ and one non-key frame $I^i$ are shown for simplicity. Key frames are firstly fed into $N_{f}$ to produce high-level features $F_{h}^{k1}$ and $F_{h}^{k2}$ while non-key frames are fed into a low-cost network $N_{l}$ to extract low-level features $F_{l}^i$. Based on the high-level features, a temporal feature $F_t$ is maintained by \textit{Recursive Feature Updating} (RFU) to model the temporal appearance of videos, where $F_t$ is updated recursively. Meanwhile, \textit{Dense Feature Transforming} (DenseFT) is utilized to propagate semantic features from updated $F_t$ from the nearest key frame to non-key frames. This whole procedure is applied along the entire sequence. PSLA is embedded in RFU and DenseFT for feature alignment and propagation. The outputs of RFU or DenseFT are fed into a task network $N_t$ to produce the detection results.}
    \label{fig2}
    \vspace{-0.5cm}
    \end{figure}

    The pipeline of our framework is illustrated in Fig. \ref{fig2}. Given a video, each frame is first processed by a CNN to extract features; it is followed by a task network $N_t$ for a specific task, such as object detection in this paper. To save the computational cost, frames are divided into key frames and non-key frames in our framework, for which the feature extraction networks are different, denoted as $N_f$ and $N_l$ respectively. The feature extraction network for non-key frames $N_l$ is a more lightweight one than $N_f$. In addition, to make use of the long term temporal information embedded in the video, a temporal feature $F_t$ is maintained across the whole video, which is gradually updated at key frames by the proposed \textit{Recursive Feature Updating} (RFU) module. Aided by the temporal feature, the semantic features of key frames will also be enhanced by RFU to benefit the final task. Meanwhile, due to the lightweight network used for non-key frames, their features are less powerful for the final task. For this reason, the \textit{Dense Feature Transforming} (DenseFT) module is proposed to enrich their features by propagating from the temporal feature $F_t$. The key assumption for such a design is that the contents of non-key frames are similar to that of nearby key frames. The core of RFU and DenseFT is to align and propagate the temporal feature to that of the currently processed frame, which is addressed by the \textit{Progressive Sparse Local Attention} (PSLA) module. In the following, we will describe in detail the proposed PSLA, RFU, and DenseFT.

\subsection{Progressive Sparse Local Attention}\label{psla}
    The core of our framework is to align and propagate feature maps across frames. To this end, we introduce Progressive Sparse Local Attention (PSLA), a novel module that aims to establish the spatial correspondence between two feature maps in order to propagate features among them.

\begin{figure}
    \includegraphics[width=0.48\textwidth]{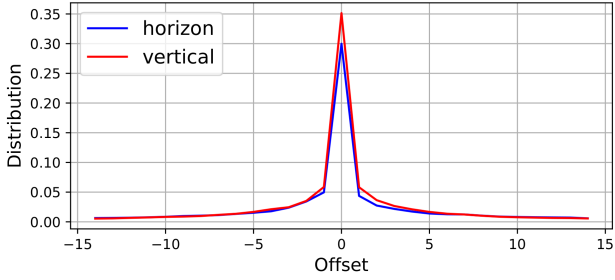}
    \caption{Optical flow field of sampled 100 ImageNet VID videos computed by FlowNet in horizon and vertical dimension. Best viewed in color.}
    \vspace{-0.2cm}
    \label{fig3}
    \end{figure}

    PSLA proceeds by first computing correspondence weights based on the feature affinities between pairs of feature cells, originating from two distinct feature maps and distributed in progressively \textit{sparser} strides (see Fig. \ref{fig1}). The motivation for this strategy is originated from Fig. \ref{fig3}, where the marginal distributions of optical flow field along the vertical and horizon axis\footnote{Precisely, the optical flow field is computed with FlowNet\cite{flownet}, on 100 videos randomly sampled from ImageNet VID training split\cite{imagenet}} are largely concentrated around zero. This suggests that the feature cells used to compute correspondence weights can be limited to a neighborhood with progressively sparser strides. This setting enables PSLA to focus more on nearby positions~(associated with small motions) and less on the positions far away~(associated with larger motions) and also in accordance with the characteristic of visual perceptional organization of retina \cite{wandell2015computational}.

    Formally, let $F^t$ and $F^{t+\epsilon}$ be feature maps of frame $I^t$ and  $I^{t+\epsilon}$ respectively, and their corresponding embedded features are denoted as $f(F^t)$ and $g(F^{t+\epsilon})\in \mathbb{R}^{c\times h\times w}$, where $c, h, w$ is the number of channels, height and width of embedded feature maps respectively. The embedding functions $f(\cdot)$ and $g(\cdot)$ here are used to reduce the channel dimension of $F^t$ and $F^{t+\epsilon}$ for saving computation. PSLA compares each feature cell from $g(F^{t+\epsilon})$ to surrounding cells from $f(F^{t})$ at local sparse locations. The resulting feature affinities are normalized to produce the weights used to align $F^{t}$. The feature cells with higher affinities, indicating higher correspondence, will get higher weights and a larger proportion of their information is propagated to a new feature cell. Finally, the aligned features are propagated to frame $I^{t+\epsilon}$. At this stage, we aim to explain the general operations of PSLA thus do not specify how $F^t$ and $F^{t+\epsilon}$ come from, which will be clarified in sect. \ref{rfu} and \ref{dft}.

    Specifically, the operation of PSLA can be formulated to two steps as follows:
    The first step is to produce sparse correspondence weights based on the feature affinities. Given two feature maps $F^t$ and $F^{t+\epsilon}$, embedded via two functions $f(\cdot)$ and $g(\cdot)$, the procedure to compute affinity between two feature cells at positions $p_1$ and $p_2$ is defined as
    \vspace{-0.1cm}
    \begin{equation}
    c_{(p_1, p_2)} = \left \langle g(F_{(x_1, y_1)}^{t+\epsilon}),f(F_{(x_2, y_2)}^{t})\right \rangle, \label{eq1}
    \vspace{-0.1cm}
    \end{equation}
    where $(x_1, y_1)$ and $(x_2, y_2)$ are position coordinates of $p_1, p_2$ respectively and $g(F_{(x_1, y_1)}^{t+\epsilon}),f(F_{(x_2, y_2)}^{t}) \in R^{c\times 1\times1}$. $\left \langle \cdot \right \rangle$ represents the inner product. For each location $(x,y)$ in $g(F^{t+\epsilon})$, only the positions in $\Phi(x,y)$ of $f(F^t)$ are considered; $\Phi(x,y)$ is a neighborhood defined by progressively sparser strides and a max displacement $d$. For clarity, we divide $\Phi(x,y)$ into a series of sub-regions as
    \begin{equation}
    \Phi(x,y) = \{\Phi_0(x,y), \Phi_1(x,y),...,\Phi_d(x,y)\},  \label{eq2}
    \end{equation}
    where
    \begin{small}
    \begin{equation}
    \vspace{-0.2cm}
    \begin{split}
    \Phi_0(x,y) &= \{(x,y)\},\\
    \Phi_s(x,y) &= \{(x+a, x+b), \forall a, b \in \{s,0,-s\}\}\backslash \{(x,y)\}, \label{eq3}
    \end{split}
    \end{equation}
    \end{small}
    $s$ is set to satisfy $1\leq s\leq d$ in our implementation. $\Phi_s(x,y)$ stands for the positions in sub-region with stride $s$. The spatial arrangement of $\Phi(x,y)$ in $g(F^t)$ is shown in Fig. \ref{fig1}, where regions of different colors correspond to different sub-regions $\Phi_s(x,y)$. As stated in the beginning of this section, it is designed as a progressively sparser grid from center to outside. Then we can compute the norma\-lized correspondence weights:
    \begin{equation}
        \hat{c}_{(p_1,p_2)} = \frac{exp(c_{(p_1,p_2)})}{\sum_{p_2 \in \Phi(x_1, y_1)}exp(c_{(p_1,p_2)})}. \label{eq4}
    \end{equation}
    By introducing a \textit{softmax} as normalization, we force the weights to compete with each other. As a result, PSLA can capture the most similar and critical feature in the region, similar to an attention mechanism \cite{bahdanau2014neural} and can implicitly establish spatial correspondence between two feature maps.

    Then, in the second step, $F^{t}$ can be aligned with $F^{t+\epsilon}$  by aggregating the corresponding feature cells with correspondence weights:
    \vspace{-0.2cm}
    \begin{equation}
    \hat{F}_{(x_1,y_1)}^{t+\epsilon} =  \sum_{p_2:(x_2, y_2)\in \Phi(x_1,y_1)} \hat{c}_{(p_1,p_2)}F_{(x_2,y_2)}^{t}. \label{eq5}
    \vspace{-0.2cm}
    \end{equation}
    The procedure of aligning feature using PSLA can be formulated as $\hat{F}^{t+\epsilon}=\textit{PLSA}(F^{t+\epsilon}, F^t)$, which is the core module embedded in RFU (sect. \ref{rfu}) and DenseFT (sect. \ref{dft}).

\subsection{Recursive Feature Updating}\label{rfu}
    Videos provide rich information that are beneficial for object recognition, \eg visual cues and temporal context from nearby frames. However, image object detectors ignore the appearance and context information from previous frames in a video sequence. This inspires us to propose \textit{Recursive Feature Updating} (RFU). RFU is a procedure that aggregates and integrates semantic features of sparse key frames along time, aiming to increase detection accuracy by exploiting temporal context.
    \begin{figure}
    \includegraphics[width=0.45\textwidth]{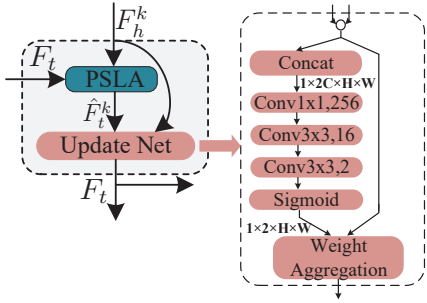}
    \caption{Recursive Feature Updating (RFU). (Conv$k\times k, n$) is a convolution layer with kernel size of $k$ and $n$ output channels.}\label{fig4}
    \vspace{-0.4cm}
    \end{figure}

    Specifically, RFU maintains and updates a temporal feature $F_{t}$ with semantic features of sparse key frames recursively throughout the whole video. In this procedure, directly updating $F_t$ with the feature of a new key frame can be problematic because the movement of objects in videos would generate misaligned spatial features.
    Therefore PSLA is exploited to enforce the spatial consistency between $F_t$ and high-level features of the new key frame. Given a high-level feature $F_h^{k}$ of a new key frame $I^{k}$ ($k$ is the index of key frame in the image sequence), the operation of PSLA can be formulated as $\hat{F}_{t}^{k}=\textit{PSLA}(F_h^{k}, F_t)$.

    After aligning the temporal feature, a tiny neural network, named \textit{Update Net}, is devised to fuse $\hat{F}_{t}^{k}$ with $F_h^{k}$ adaptively, with the goal of incorporating temporal context of videos into $\hat{F}_{t}^{k}$. As shown in Fig.~\ref{fig4}, \textit{Update Net} takes the concatenation of $\hat{F}_{t}^{k}$ and $F_{h}^{k}$ as inputs. Then it produces the adaptive weights $\hat{W}^{k}$ and $W^{k}$ through multiple layers of convolution, where $\hat{W}^{k}$ and $W^{k}\in \mathbb{R}^{1\times h\times w}$ indicate the importance of feature cells at each spatial location of two different feature maps. The weights are norma\-lized over two feature maps for every spatial location so that $\hat{W}_{ij}^{k}+W_{ij}^{k}=1$. Finally, $F_{t}$ is updated based on the weights:
    \vspace{-0.2cm}
    \begin{equation}
    F_{t} = \hat{W}^{k} \cdot \hat{F}_{t}^{k} + W^{k} \cdot F_{h}^{k},
    \vspace{-0.2cm}
    \end{equation}
    where $\cdot$ is the Hadamard product (\ie element-wise multiplication) after broadcasting the weight maps. Finally the updated $F_t$ is used in place of $F_h^{k}$ to produce results of key frame $I^{k}$ and taken as the updated temporal feature.

\subsection{Dense Feature Transforming}\label{dft}

    Since the features extracted by $N_l$ for non-key frames are less powerful, we introduce Dense Feature Transforming (DenseFT) to generate semantic features of non-key frames by feature transformation and propagating from the maintained temporal feature $F_t$.

    Specifically, the extracted low-level features $F_l$ are used by PSLA to propagate semantic features from the temporal feature $F_t$ at the nearest key frame. However, these low-level features do not contain sufficient semantic information to find spatial correspondence. The aligned feature may fail to preserve critical information. To address this issue, a light-weight network \textit{Transform Net} is employed to further encode the extracted low-level features, aiming to approximate the high-level semantic features. This is a pivotal step because it not only enriches the semantic information of low-level features but also avoids the gradients gene\-rated by feature propagation flowing into $N_l$ directly, hence improving the robustness of training as well. The encoded features are fed into PSLA to align $F_t$ with non-key frames.
    \begin{figure}
    \includegraphics[width=0.45\textwidth]{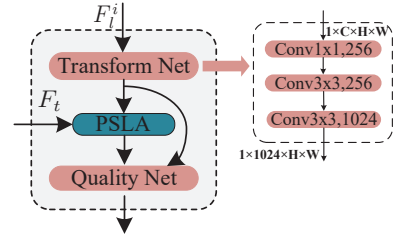}
    \caption{Dense Feature Transforming (DenseFT).} \label{fig5}
    \vspace{-0.5cm}
    \end{figure}

    After propagating $F_t$ to non-key frames, we fuse it with the low-level features $F_l$. The reason for this is the aliasing effect, caused by weighted aggregation in feature alignment, may make the propagated feature lose some details of object appearance that are important for recognition. To this end, a network \textit{Quality Net} is embedded in DenseFT to complement detail information. Finally, the output of \textit{Quality Net} is fed into $N_t$ to produce results of non-key frames.

\vspace{-0.2cm}
\subsection{Implementation Details}
    We use ResNet-101 pretrained on ImageNet as $N_f$ for feature extraction, whose layers lower than \textit{res4b3} (including \textit{res4b3}) are selected to construct $N_{l}$. Following \cite{dff}, an RPN is used to generate region proposals and R-FCN is used as the task-specific network $N_t$ for object detection.  The embedding function $f(\cdot)$ and $g(\cdot)$ in Equ. (\ref{eq1}) are implemented with $1\times1$ convolution layers with 256 filters. For the hyper-parameter of PSLA, max displacement $d$ is set as 4 by default. The whole network including RFU and DenseFT is trained end-to-end on 8 GPUs for 120K iterations using SGD. Learning rate is $2.5\times 10^{-4}$ in the first 80K iterations and $2.5\times 10^{-5}$ in the last 40K iterations. During testing, we employ a fixed key frame schedule simi\-lar to \cite{dff}, \ie,  a video is split into segments containing an equal number of frames and the middle frames are chosen as key frames. Key frame interval $l$ is set as 10 by default.

    The details of \textit{Update Net} is illustrated in Fig.~~\ref{fig4}. A $1\times1$ convolution layer is first used to reduce the features to 256 channels, which is followed by two $3\times3$ convolution layers with 16 and 2 filters respectively, to produce the corresponding spatial weights for each feature. The structure of \textit{Quality Net} is the same as \textit{Update Net}.

    As shown in Fig.~\ref{fig5}, \textit{Transform Net} is implemented with a bottleneck block. Firstly, a convolution layer with $1\times1$ kernel is used to reduce the feature channels. Then, two successive $3\times3$ convolution layers with 256 and 1024 filters respectively are appended to further encode the feature.

\section{Experiments}
\subsection{Dataset and Setup}
    We evaluate our framework on ImageNet VID \cite{imagenet} dataset that contains objects of 30 classes with fully annotated bounding boxes. Following the protocols in \cite{dff}, the model is trained on the intersection of training split of VID and a subset of ImageNet DET \cite{imagenet} with the same categories as VID. The trained model is tested on the validation split of VID.

    During training, we train the network with a batch of three images on each GPU. Each batch is sampled from either ImageNet VID or ImageNet DET at $1:1$ ratio. When sampling from VID, we first sample an image as a non-key frame $I^{i}$. Then we sample another two images $I^{k1}$ and $I^{k2}$ near the non-key frame in a random offset as key frames. Specifically, if $I^{i}$ is the $n^{th}$ frame of a video then $I^{k1}$ lies in $[-l+n,-0.5l+n]$ and $I^{k2}$ lies in $[-0.5l+n,0.5l+n]$, where $l$ is the key frame interval. Three images are the same when sampling from ImageNet DET. Only non-key frames are provided with labels in the training phase.

\subsection{Results}

    We compare our framework with several state-of-the-art methods for video object detection w.r.t accuracy and complexity.

    The results are shown in Table~\ref{tab1}. Our method achieves 77.1\% mAP at runtime of 30.8$\backslash$18.7 fps on TITAN V$\backslash$X when using ResNet101 as a backbone. It surpasses the frame baseline~(\ie R-FCN~\cite{rfcn}) both in terms of accuracy and runtime, showing the potential of exploiting temporal information in videos to improve the object detection performance. Compared to the optical flow based methods such as DFF~\cite{dff} and FGFA~\cite{fgfa}, our method achieves much higher mAP and is only slightly slower than DFF. However, it is worth to note that our method significantly reduces the model parameters by nearly 34\%$(96.6M\rightarrow63.7M)$. Although MANet~\cite{fully} achieves 1\% higher mAP than our method, it is much slower. Towards~\cite{towards} is inferior to our method when using the same backbone.

    \begin{table}
    \centering
    \footnotesize
    \resizebox{0.475\textwidth}{!}{
    \begin{tabular}{ccccc}
    \hline
    \makecell[c]{Methods}                    &   \makecell[c]{mAP\\($\%$)}   &   \makecell[c]{runtime\\(fps)} &   \makecell[c]{model size\\ (params)}  & Backbone\\
    \hline
    TCN \cite{tcn}                            &     47.5                   &       -                       &      -                               &  GoogLeNet \\
    TPN \cite{tpn}                       &     68.4                   &       2.1(X)                  &      -                               &  GoogLeNet    \\
    R-FCN  \cite{rfcn}                        &     73.9                   &       4.05(K)                  &      59.6M                           &  ResNet101 \\
    TCNN \cite{tcnn}                          &     73.8                   &       -                       &      -                                &GoogLeNet \\
    DFF \cite{dff}                            &     73.1                   &         20.25(K)               &      96.6M                             &  ResNet101 \\
    D($\&$T loss) \cite{DT}                          &     75.8                   &       -                  &      -                                &  ResNet101 \\
    FGFA \cite{fgfa}                          &     76.3                   &       1.36(K)                 &      100.4M                            &  ResNet101 \\
    D$\&$T(online) \cite{DT}               &     78.7                   &        5.3(X)                &      -                                 &  ResNet101 \\
    D$\&$T($\delta=1$) \cite{DT}               &     79.8                   &       -                  &      -                                 &  ResNet101 \\
    MANET \cite{fully}                        &     78.1                   &       5(XP)                   &      -                                &  ResNet101  \\
    ST-lattice \cite{lattice}                 &     79.6                   &
          20(X)                      &      $>$ 100M                                &  ResNet101  \\
    Towards \cite{towards}                    &     78.6                   &      13.0(X)                  &      -                                 &  ResNet101+DCN  \\
    Ours                                      &     77.1                   &      30.8(V)$\backslash$18.73(X)       &       63.7M   &         ResNet101 \\
    Ours                                      &     \bf{80.0}         &      \bf{26.0(V)$\backslash$13.34(X)}          &      \bf{72.2M}                   &  ResNet101+DCN  \\
    \hline
    FGFA \cite{fgfa} + \cite{seqnms}                          &     78.4                  &       1.14(K)                 &      100.4M                            &  ResNet101 \\
    MANET \cite{fully} + \cite{seqnms}         &     80.3                   &       -                   &      -                                 &  ResNet101 \\
    STMN \cite{stmn} + \cite{seqnms}          &     80.5                  &   1.2(X)                  &      -                                 &  ResNet101 \\
    Ours + \cite{seqnms}                      &     78.6                  &   5.7(X)                  &      -                                 &   ResNet101     \\
    Ours + \cite{seqnms}                      &     \bf{81.4}                   &  \bf{6.31(V)}$\backslash$\bf{5.13(X)} &      \bf{72.2M}                             &  ResNet101+DCN \\
    \hline
    \end{tabular}
     }
    \caption{\footnotesize Performance of our method and state-of-the-art methods on ImageNet VID. Results of other methods are obtained from their papers, where different GPUs were used. X means TITAN X, XP means TITAN XP, K means K40, Ti means 1080 Ti and V means TITAN V.}
    \label{tab1}
    \vspace{-0.2cm}
    \end{table}

    Note that, due to the high efficiency of the proposed framework, it can be used with more powerful backbones to further improve the accuracy while still maintaining fast runtime. As can be seen, using ResNet101+DCN as the backbone, our method achieves 80.0\% mAP at runtime of 26.0$\backslash$13.34 fps on TITAN V$\backslash$X, which is better than recent advances in terms of both accuracy and speed. The most competing method to ours is ST-lattice~\cite{lattice}, which obtains higher fps with lower mAP. Nevertheless, ST-lattice requires an extra ResNet-101 based classifier to re-score the bounding boxes and two ResNet18 models to propagate and refine bounding boxes. For these reasons, it takes at least 100M parameters (required by all models). For compa\-rison, our best model is much smaller, requiring about 72M parameters.

    After combining with Seq-NMS \cite{seqnms}, the mAP of our method finally reaches $81.4\%$, outperforming all the state-of-the-art methods to the best of our knowledge. MANet~\cite{fully} and STMN~\cite{stmn} also achieve very high mAPs when combined with Seq-NMS, however, they suffer from high computation complexity since they use more than 10 nearby frames to augment the feature of the reference frame. Conversely, our method only requires few key frames to propagate the features and in the meantime reduces the feature extraction time for non-key frames with a lightweight network, thus significantly reducing its runtime. To sum up, the overall performance of our method is better than previous works, achieving a very good tradeoff among accuracy, speed and model size.

\subsection{Ablation Study}\label{ablation}

    We conduct ablation study on ImageNet VID to validate the effectiveness of PSLA and the proposed framework. After introducing different configurations used for ablation study, we firstly compare PSLA to existing non-optical flow alternatives for feature propagation. Then we compare PSLA to optical flow. Finally, we conduct ablation study on different modules of the proposed framework. We also show that the proposed framework is general enough to benefit other kinds of feature propagation methods.

   \begin{table}
    \resizebox{0.475\textwidth}{!}{
    \begin{tabular}{c|c|c|c|c}
    \hline
    \makecell[cc]{Methods}    &    \makecell[cc]{Feature\\Propagation}    &   \makecell[cc]{Transform \\Net}  &  \makecell[cc]{RFU}      &   \makecell[cc]{Quality\\ Net}\\
    \hline
    Nonlocal\_S               &                Nonlocal \cite{nonlocal}                  &       \cmark                      &          \xmark          &   \xmark   \\
    Nonlocal\_F               &                Nonlocal \cite{nonlocal}                 &       \cmark                      &          \cmark          &   \cmark   \\
    \hline
    MatchTrans\_S               &                MatchTrans \cite{stmn}                  &       \cmark                      &          \xmark          &   \xmark   \\
    MatchTrans\_F               &                MatchTrans \cite{stmn}                 &       \cmark                      &          \cmark          &   \cmark   \\
    \hline
    DensePSLA\_S               &                Dense PSLA                  &       \cmark                      &          \xmark          &   \xmark   \\
    DensePSLA\_F               &                Dense PSLA                  &       \cmark                      &          \cmark          &   \cmark   \\
    \hline
    our method (a)           &                 PSLA                       &            \cmark                 &         \xmark            &  \xmark        \\
    our method (b)           &                  PSLA                      &              \cmark               &         \cmark            &   \xmark        \\
    our method (c)            &                  PSLA                     &            \cmark                 &         \cmark          &   \cmark   \\
    \hline
    \end{tabular}
    }
    \caption{The configuration of different methods for ablation study.}
    \label{tab2}
    \vspace{-0.2cm}
    \end{table}

    Besides relying on the widely used optical flow to propagate feature maps, there are two typical alternatives in literature, MatchTrans~\cite{stmn} and Nonlocal~\cite{nonlocal}. Basically, MatchTrans computes the propagation weights by accumulating all similarity scores in the local region while Nonlocal considers all positions. By contrast, PSLA uses a progressively sparse local region. It also applies softmax when computing the propagation weights so that spatial correspondence can be implicitly established. To better analyse the effectiveness of using progressively sparse local region, we implement a dense version of PSLA (denoted as DensePSLA) which uses all positions in the local region as MatchTrans does but with the same way to compute propagation weights as PSLA does~(\ie with softmax, Equ.~(\ref{eq4})). Furthermore, to show the performance of different feature propagation methods, a simple object detection framework is implemented by only propagating the feature of preceding key frame to non-key frames. These methods are denoted by adding \_S to the propagation methods, such as Nonlocal\_S, MatchTrans\_S. By contrast, \_F means using the RFU and DenseFT modules in our video detection framework. All these evaluated methods in ablation study are summarized in Table~\ref{tab2}. Note that the \textit{Transform Net} is used on all these methods as it enables stable training according to our experiments.

    \vspace{-0.35cm}
    \paragraph{Performance of different feature propagation methods} The results of using different feature propagation methods are listed in Table.~\ref{tab3}. By attending to the local region instead of all positions, our method(a) outperforms Nonlocal\_S by a large margin. Moreover, when comparing to MatchTrans\_S and DensePSLA\_S, our method(a) achieves better results at all max displacement settings and consumes less runtime as well, demonstrating the effectiveness and importance of introducing progressive sparsity in PSLA.

    \begin{table}
    \centering
    \resizebox{0.475\textwidth}{!}{
    \begin{tabular}{lcccc}
    \hline
    \makecell[cc]{Methods}                  &   \makecell[cc]{max\\displacement}  &        \makecell[cc]{mAP\\($\%$)}   &   \makecell[cc]{runtime\\(fps)}  & \makecell[cc]{parameters\\(M)}\\
    \hline
    Nonlocal\_S              &           -          &        72.1        &      40(V)          & 62.7\\
    \hline
    MatchTrans\_S       &           2          &          71.4         &       41.2(V)          & 62.7 \\
    DensePSLA\_S        &           2          &         72.9       &       41.2(V)    & 62.7 \\
    our method (a)       &           2         &           \bf{73.6}     &       \bf{42.7(V)}    & 62.7 \\
    \hline
    MatchTrans\_S       &           3          &          71.5         &       40.8(V)         & 62.7\\
    DensePSLA\_S        &           3          &         73.7       &       40.8(V)    & 62.7\\
    our method (a)       &           3          &          \bf{74.3}      &       \bf{42.5(V)}    &  62.7 \\
    \hline
    MatchTrans\_S       &           4          &          72.5         &        40.6(V)         & 62.7 \\
    DensePSLA\_S        &           4          &         73.6       &       40.6(V)    & 62.7 \\
    our method (a)       &           4          &          \bf{74.4}      &       \bf{42.0(V)}    & 62.7 \\
    \hline
    MatchTrans\_S       &           5          &          72.4         &        40.2(V)         &   62.7\\
    DensePSLA\_S        &           5          &          73.0      &       40.2(V)    &   62.7\\
    our method (a)       &           5         &          \bf{73.8}      &       \bf{41.4(V)}    &    62.7\\
    \hline
    \end{tabular}}
    \caption{Comparison of different feature propagation methods.}
    \label{tab3}
    \vspace{-0.2cm}
    \end{table}

    \begin{table}
    \centering
    \begin{tabular}{lccc}
    \hline
    \makecell[cc]{Methods}            &   \makecell[cc]{max\\displacement}   &   \makecell[cc]{key frame\\interval} & \makecell[cc]{mAP($\%$)} \\
    \hline
    DFF*       &   -                   &     15        &  72.2          \\
    DFF*       &   -                   &     25        &  69.7         \\
    DFF*       &    -                  &     35        &  67.5         \\
    \hline
    our method (a)      &  4            &    15          &   72.9 (+0.7)      \\
    our method (a)      &  4            &    25          &   70.5 (+0.8)     \\
    our method (a)      &  4            &    35          &   68.5 (+1.0)     \\
    \hline
    \end{tabular}
    \caption{Comparison between PSLA and DFF at different key frame intervals. * means our re-implementation.}
    \label{tab4}
    \vspace{-0.2cm}
    \end{table}

    \begin{figure*}
    \centering
    \includegraphics[width=1.0\textwidth]{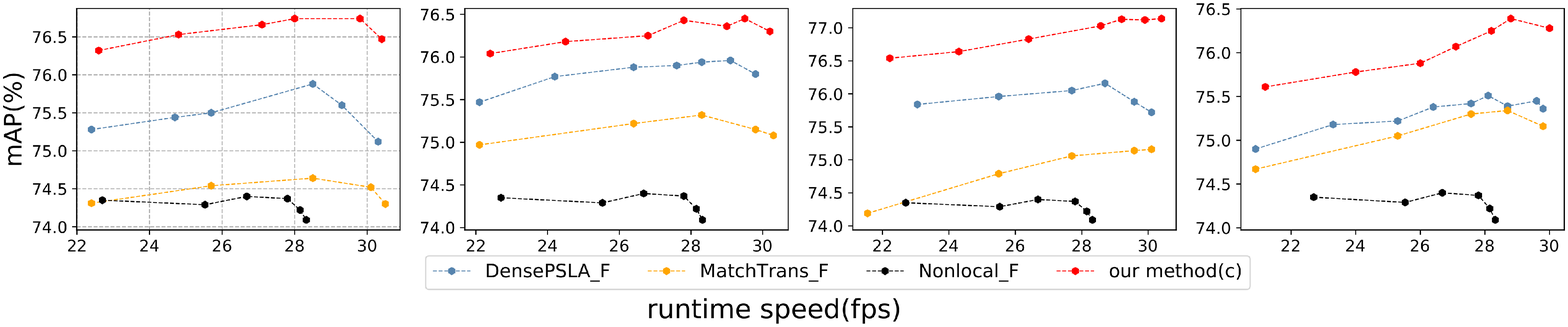}
    \caption{mAP vs. runtime for different methods. The results from left to right correspond to max displacement $d$, from 2 to 5.}\label{fig6}
    \vspace{-0.45cm}
    \end{figure*}

    Fig.~\ref{fig6} shows the trade-off between speed and accuracy of different methods in our video detection framework with different max displacements\footnote{Nonlocal is a global method, so it does not have a parameter of max displacement. Thus, the four curves of Nonlocal\_F is identical in Fig.~\ref{fig4}}. Essentially, larger key frame interval means a larger number of non-key frames whose feature extraction is significantly reduced, thus faster running time will it be. Therefore, by setting different key frame intervals for different methods, we can obtain different mAPs versus different speeds. It is obvious that the proposed method consistently outperforms other competitive methods at all evaluated key frame intervals. We can also observe from Fig.~\ref{fig6} that, mAP increases along with the speed at the beginning~(\ie, small key frame interval) but decreases when the key frame interval reaches a large number. On one hand, small key frame interval only causes small motion of objects between the key and non-key frames which is hard to be captured in high-level feature maps whose receptive field is 16$\times$16. Thus, feature propagation may aggregate harmful information and hurt the performance. On the other hand, too large key frame interval leads to a very large motion of objects, in which case establishing spatial correspondence is quite challenging. As a result, the accuracy decreases when the key frame interval is set either too small or too large.

    \vspace{-0.4cm}
    \paragraph{PSLA VS. optical flow} In order to validate the advantage of PSLA on capturing spatial correspondence on feature maps, we compare PSLA with DFF~\cite{dff}, a pioneer work on video object detection with optical flow. The results are illustrated in Table \ref{tab4}, where the results of DFF are obtained by our own implementation. Obviously, our method(a) performs much better than DFF. The larger the key frame interval is, the more significant the relative improvement is. It verifies that by directly establishing the spatial correspondence in feature maps, PSLA aligns two feature maps better than aligning them based on pixel-level correspondence from optical flow.

    \begin{table}
    \centering
    \begin{tabular}{lcc}
    \hline
    Methods    & mAP($\%$)           &  runtime (fps)   \\
    \hline
    our method(a)                                 &    74.4         &      42.0       \\
    our method(b)                             &     75.8        &     31.2        \\
    our method(c)                             &      \bf{77.1}       &     30.8     \\
    \hline
    Nonlocal\_S                               &       72.1      &    40.0           \\
    Nonlocal\_F                                &       74.1      &     28.3           \\
    \hline
    MatchTrans\_S                             &        72.5     &     40.6                 \\
    MatchTrans\_F                     &        75.2        &   30.1                           \\
    \hline
    DensePSLA\_S                      &       73.6      &       40.6                         \\
    DensePSLA\_F                      &       75.7      &        30.1                        \\
    \hline
    \end{tabular}
    \caption{The proposed video detection framework benefits various feature propagation methods. The max displacement is set as 4. All results are tested on TITAN V.}
    \label{tab5}
    \vspace{-0.2cm}
    \end{table}

    \vspace{-0.2cm}
    \paragraph{Effectiveness of the proposed framework} Table~\ref{tab5} gives the results of our method when gradually adding RFU and DenseFT modules. Firstly, only using PSLA to propagate features from the nearby key frame achieves 74.4\% mAP. Then, by exploiting RFU to model the temporal appearance used for feature propagation~(\ie, method(b)), the performance of our method(a) is improved by $1.4\%$. Finally, the result is further improved to $77.1\%$ by adding DenseFT to enhance the feature representations of non-key frames. RFU and DenseFT have also been used in other feature propagation methods, and we can observe consistent performance improvement from Table~\ref{tab5}.


\subsection{Extension to other tasks}
    The proposed framework shown in Fig.~\ref{fig2} could be actually used for other vision tasks beyond object detection studied in this paper. Here we conduct a simple experiment on video object segmentation to demonstrate the possible extension of our method. Specifically, we replace the R-FCN used in this paper with deeplab \cite{deeplabv2} for semantic segmentation in videos. In this case, our method is similar to Low-latency \cite{lowlatency}. The difference is that Low-latency predicts location-adaptive kernel weights to generate features of non-key frames while we utilize the parameter-free PSLA to perform feature alignment. The experiment is conducted on CityScapes \cite{cityscapes}, which contains snippets of street scenes from 50 different cities. We train our framework on the training set and evaluate the pixel-level mean intersection-over-union (mIoU) on the validation set. More training details are given in supplementary materials.

    \begin{table}
    \centering
    \begin{tabular}{lcc}
    \hline
    Methods                 &     mIoU($\%$)   &   runtime (fps) \\ \hline
    DVS \cite{dynamic}      &      70.4        &        19.8(Ti)     \\
    DFF \cite{dff}          &      69.2        &        5.6(K)     \\
    Frame baseline \cite{dff} &     71.1    &         1.52(K)    \\
    \hline
    DFF (*)          &      69.8        &        15.4(V)       \\
    Frame baseline (*)          &      72.1        &        6.2(V)       \\
    \hline
    Ours                    &      71.9        &        11.6(V)        \\
    \hline
    \end{tabular}
    \caption{Comparison of different methods on Cityscape. * means our re-implementation.}
    \label{tab6}
    \vspace{-0.5cm}
    \end{table}

    The results are summarized in Table \ref{tab6}. For a fair comparison, we re-implemented the frame baseline and DFF with the same setting as our method. Our method achieves very close performance to the frame baseline with higher fps, verifying the effectiveness of the proposed PSLA. It also achieves the best mIoU comparing to DFF and DVS at a reasonable speed. For comparison, DFF achieves faster runtime than the baseline at the cost of accuracy by a large margin. Comparing to DVS \cite{dynamic}, a better mIoU is achieved by ours. Although faster, DVS relies on the FlowNet to propagate features, thus having much more parameters and being less preferable to practical scenarios. As for Low-latency, it is hard to compare directly since its segmentation head is not specified. Our good result on video semantic segmentation demonstrates the universality of the proposed method for video recognition.

\vspace{-0.2cm}
\section{Conclusion}
    In this paper, we propose a novel framework for video object detection. At its core, a novel module PSLA is proposed to propagate features effectively. In addition, two techniques RFU and DenseFT are designed to model temporal appearance and enhance feature representations. We conduct ablation studies on ImageNet VID to prove the effectiveness of our framework on video object detection. The proposed framework achieves 81.4$\%$ mAP on ImageNet VID and outperforms state-of-the-art methods. Additional experiment of video semantic segmentation on CityScapes demonstrates the generalization ability of the framework.

\vspace{-0.3cm}
\section*{Acknowledgement}

This work was supported by the National Natural Science Foundation of China under Grants 61573352, 61876180, 91646207, 61773377, the Young Elite Scientists Sponsorship Program by CAST (2018QNRC001), and the Beijing Natural Science Foundation under Grant L172053.

{\small
\bibliographystyle{ieee_fullname}
\bibliography{egbib}
}

\end{document}